\documentclass{article}

\usepackage[T1]{fontenc}
\usepackage[utf8]{inputenc}
\usepackage{microtype}
\usepackage{graphicx}
\usepackage{booktabs}
\usepackage{amsmath,amssymb}
\allowdisplaybreaks
\usepackage{array,tabularx}

\usepackage{hyperref}

\ifdefined\fublind
  \usepackage{mlsys2025}
  \ifdefined\fulinenumbers\else\ClearShipoutPicture\fi
\else
  \usepackage[accepted]{mlsys2025}
  \makeatletter
  \renewcommand{\printAffiliationsAndNotice}[1]{%
    {\let\thefootnote\relax\footnotetext{\hspace*{-\footnotesep}#1%
    Correspondence to: \mlsyscorrespondingauthor@text.}}}
  \makeatother
\fi

\usepackage{xcolor,tikz,pgfplots}
\usetikzlibrary{arrows.meta,positioning,calc}
\pgfplotsset{compat=1.18}

\definecolor{ink}{HTML}{183A5C}
\definecolor{teal}{HTML}{087F8C}
\definecolor{orange}{HTML}{C45D21}

\newcommand{\fu}{\textsc{FuTransformer}}
\newcommand{\code}[1]{\texttt{#1}}
\newcommand{\speed}{\ensuremath{\times}}

\newcolumntype{Y}{>{\raggedright\arraybackslash}X}

\mlsystitlerunning{FuTransformer: FP8 Prefill and INT4 Decode for Llama-Style Inference on Hopper}

\begin{document}

\twocolumn[
\mlsystitle{FuTransformer: FP8 Prefill and INT4 Decode for Llama-Style Inference on Hopper}

\begin{mlsysauthorlist}
\mlsysauthor{Wei Wang}{}
\ifdefined\fublind\else\par\vspace{2pt}\href{mailto:wei.wang217@gmail.com}{\texttt{wei.wang217@gmail.com}}\par\fi
\end{mlsysauthorlist}

\ifdefined\fublind\mlsysaffiliation{anon}{}\fi  
\mlsyscorrespondingauthor{Wei Wang}{wei.wang217@gmail.com}

\mlsyskeywords{LLM inference, Transformer, CUDA, Hopper, FP8, INT4, KV cache, quantization}

\vskip 0.3in

\begin{abstract}
Transformer inference has different bottlenecks during prefill and decode. Prefill reuses weights across many tokens and is dominated by matrix multiplication and activation data movement. Single-token decode has limited data reuse and is dominated by weight and KV-cache reads. We present \fu, a CUDA implementation of a Llama-style Transformer block for NVIDIA Hopper GPUs. CODA showed how to express Transformer training operators as GEMM-plus-epilogue programs. \fu\ applies this principle to inference. It adds direct KV-cache placement, folds the fixed RMSNorm gain into weights, writes FP8 values directly from the producing epilogue using delayed scaling, and supports prefill without padding. For decode, \fu\ uses INT4 projection weights and an INT4 copy of the KV cache. Keys and weights follow the Tensor Core fragment layout, so the kernels can dequantize them directly into MMA operands without creating full intermediate tiles. A block-diagonal Hadamard transform makes INT4 cache quantization less sensitive to outlier channels. On one H100, prefill is up to 1.48\speed\ faster than compiled PyTorch FP8 and reaches 1.29 PFLOP/s. For the closest shared operation, gate/up projection followed by SwiGLU, the BF16 path is 1.13\speed\ faster than the released CODA kernel at the 8B shape. Our FP8 inference path is 2.02--2.18\speed\ faster than the evaluated BF16 CODA path. Across the 8B and 70B shapes at batch 1 to 32, INT4 decode is 1.3--2.4\speed\ faster than whichever PyTorch decode block is faster at each point: 2.1--3.3\speed\ faster than the compiled, CUDA-graphed BF16 block, and 1.3--6.3\speed\ faster than the same block using PyTorch's INT4 projection kernel, which is built for a single query row and loses to its own BF16 block at batch 32. INT4 decode is also 31--51\% faster than our FP8 decode path. For the complete Llama-3.1-8B model, one INT4 decode step takes 2.8~ms at batch one, compared with 8.1~ms for the compiled transformers implementation. On Llama-3.1-8B, decoding every WikiText-2 token through the INT4 weight path increases perplexity from 6.24 to 7.23 with round-to-nearest weights, or to 6.99 with GPTQ weights stored in the same format. The rotated INT4 cache adds another 0.20. The kernel path matches the quantized reference computation to within 0.001 perplexity.
\end{abstract}
]

\printAffiliationsAndNotice{}

\section{Introduction}

Matrix multiplication dominates Transformer FLOPs, but data movement and other operations also contribute to execution time. A decoder block also performs normalization, residual updates, position encoding, activation functions, precision conversion, and KV-cache updates. These operations perform little arithmetic but often read and write large intermediate tensors. As Tensor Cores become faster, this data movement takes a larger share of the total execution time.

CODA~\cite{coda} addresses this problem for training by expressing many Transformer operators as GEMM-plus-epilogue programs. An optimized GEMM mainloop produces an output tile. A programmable epilogue then processes the tile while it is still on chip. This allows pairwise operations such as RoPE and SwiGLU, as well as residual updates, scaling, and partial reductions, to run without separate reads and writes to global memory.

\fu\ uses the same GEMM-plus-epilogue principle for autoregressive inference. Inference adds two requirements. First, prefill must manage persistent state. It must write K and V directly to the KV cache, produce low-precision activations for later projections, and support requests of different lengths. Second, decode has too few rows to behave like a large GEMM. At small batch sizes, each layer reads its weights for only a few rows, and attention reads cached K and V values for every query. The decode path must therefore reduce the bytes read and prevent dequantization from delaying the main computation.

For prefill, \fu\ uses persistent FP8 GEMMs with epilogue policies selected at compile time. These policies apply RoPE or SwiGLU, write K and V to the cache, collect RMSNorm statistics, restore scales, and optionally write FP8 output directly. Inference weights are fixed, so the learned RMSNorm gain can be folded into the next projection before quantization. The row-wise normalization factor is then applied in that projection's epilogue. This removes the normalized activation tensor while preserving the quantized computation. Packed inputs also remove padded projection rows from batches with different request lengths.

For decode, \fu\ uses INT4 weights and an INT4 copy of the KV cache. Hopper has no INT4 Tensor Core MMA, and its FP8 MMA requires large warpgroup shapes that the few query rows in decode cannot fill. The kernel must therefore expand INT4 values before a 16-bit MMA. The main design choice is where to perform this expansion. \fu\ stores keys and weights in the order required by the MMA fragments. A byte-indexed lookup converts packed cache values into FP16 pairs. A short sequence of logic and packed BF16 operations converts weight nibbles directly in registers. Group scales are applied after accumulation. Before cache quantization, a block-diagonal Hadamard transform spreads large channel values. Matching transforms on the query and attention output preserve the attention computation before quantization.

The main contributions are:

\begin{itemize}
\item \textbf{Inference-specific GEMM epilogues.} Starting from CODA's GEMM-plus-epilogue formulation, \fu\ adds direct KV-cache placement, folds the fixed RMSNorm gain into weights, writes FP8 producer output with delayed scaling, and supports packed prefill.
\item \textbf{Fragment-aligned INT4 decode.} INT4 weights and cache entries are packed in the layout required by the MMA fragments that consume them. Values are dequantized in registers or through a small lookup table. Scales are applied outside the per-element inner loop.
\item \textbf{Outlier-aware cache quantization.} A block-diagonal Hadamard transform reduces INT4 cache error when keys contain large channel outliers. Before quantization, the transform preserves the exact attention dot product.
\item \textbf{A final-result evaluation.} We compare full prefill and decode layers with PyTorch. We also compare the complete Llama-3.1-8B model with PyTorch and the closest shared gate/up--SwiGLU stage with CODA. We report performance and numerical error, and label comparisons that use different precisions. Using a trained Llama-3.1-8B checkpoint, we decode every WikiText-2 token through the INT4 path and measure perplexity, per-layer error, and error accumulated across layers.
\end{itemize}

\section{Background and Positioning}

\subsection{Llama-style inference block}

Let $X\in\mathbb{R}^{M\times H}$ contain $M$ token vectors, each with hidden width $H$. Let $I$ denote the feed-forward width, $H_q$ the number of query heads, $H_{kv}$ the number of KV heads, and $D$ the head dimension. We evaluate this block:

\begin{align}
\widehat X &= \operatorname{RMSNorm}(X;\gamma_1), \nonumber\\
(Q,K,V)&=\widehat XW_{qkv}^{\top}, \label{eq:block1}\\
O&=\operatorname{Attention}(\operatorname{RoPE}(Q),K_c,V_c), \nonumber\\
X_1&=X+OW_o^{\top}, \label{eq:block2}\\
\widehat X_1&=\operatorname{RMSNorm}(X_1;\gamma_2), \nonumber\\
Z&=\operatorname{SiLU}(\widehat X_1W_g^{\top})\odot(\widehat X_1W_u^{\top}), \label{eq:block3}\\
Y&=X_1+ZW_d^{\top}. \label{eq:block4}
\end{align}

$K_c$ and $V_c$ store the keys and values for earlier tokens and the current token. The block uses causal grouped-query attention (GQA)~\cite{gqa} and RMSNorm~\cite{rmsnorm}. The gate and up projections share one GEMM, giving four projection GEMMs and one attention stage in total.

During prefill, $M=BS$, where $B$ is the batch size and $S$ is the sequence length. Each weight tile is reused across many token rows. During single-token decode, $M=B$. At the small batch sizes studied here, reading the weights accounts for most projection latency. Attention latency grows with context length $T$ because attention must read the KV cache.

\subsection{GEMM-plus-epilogue programs and CODA}

A high-performance GEMM has two parts. The mainloop loads input tiles and performs matrix multiply-accumulate operations. The epilogue processes and stores each completed output tile. At that point, the output is already in registers or shared memory. The epilogue can therefore perform local operations without first writing an intermediate tensor to HBM.

CODA~\cite{coda} provides a structured way to use this idea in Transformer training. It keeps the GEMM mainloop fixed and offers five groups of epilogue operations that can be combined: elementwise and pairwise maps; vector loads and stores; tile loads and stores; tile reductions, with partial results combined by a small auxiliary kernel; and stateful transforms. CODA also shows that the row-wise RMSNorm factor can be moved through the following GEMM and applied in its epilogue. Together, these operations cover nearly all non-attention work in the forward and backward passes of a Transformer block.

FlashAttention introduced tiled attention designed to reduce data movement, together with online softmax. Later work adapted the execution schedule to newer GPU pipelines and low precision~\cite{fa1,fa2,fa3}. \fu's prefill attention follows this approach. Its decode design focuses on how low-bit K and V values are stored and expanded into the MMA fragments that use them.

\subsection{Low-precision weights and KV caches}

FP8 E4M3 reduces the data read and written by projections while retaining Tensor Core execution~\cite{fp8}. If $A_{mk}\approx a_m A^q_{mk}$ and $W_{nk}\approx b_n W^q_{nk}$, then

\begin{equation}
\widetilde C_{mn}=a_m b_n\sum_k A^q_{mk}W^q_{nk}. \label{eq:fp8_factor}
\end{equation}

The row and column scales can be applied after accumulation, rather than to every product. Systems such as Transformer Engine use delayed scaling~\cite{te}: the producer writes FP8 output immediately, using a scale collected during an earlier execution.

GPTQ and AWQ made group-scaled INT4 weights practical~\cite{gptq,awq}. Marlin showed the benefits of packing weights in Tensor Core fragment order and dequantizing them in registers~\cite{marlin}. KIVI studied low-bit KV caches~\cite{kivi}. QuaRot and SAW-INT4 use Hadamard rotations to reduce activation and cache outliers~\cite{quarot,sawint4}. \fu\ combines these ideas in a complete Hopper decode block, with layouts designed for its attention and projection fragments.

\section{FuTransformer Design}

\subsection{Overview}

\fu\ uses separate prefill and decode paths because they have different bottlenecks. Prefill keeps large GEMMs efficient and reduces data movement between them. Decode reduces the bytes read for weights and the cache, then dequantizes values in the layout required by the MMA operands.

The prefill projection kernel uses $128\times256\times128$ CTA tiles, a four-stage TMA pipeline, one producer warpgroup, and two consumer warpgroups. Its mainloop uses Hopper FP8 WGMMA and the Tensor Memory Accelerator~\cite{ptx}. An epilogue policy selected at compile time controls how each completed fragment is processed and stored. The QKV, output, gate/up, and down projections can therefore perform different inference operations while sharing one optimized GEMM mainloop.

The decode path has two kernel families. A register-streaming kernel handles $M=1$, and a skinny matrix kernel handles $2\le M\le32$. Both read packed INT4 weights and expand them directly into MMA fragments for accumulation with BF16 activations. Decode attention reads the INT4 cache copy. For long contexts, it uses split-KV reduction.

\subsection{Inference-aware FP8 prefill}

\subsubsection{Direct KV-cache placement}

In training, K and V are activations used later in the same pass. In inference, they must be written to persistent cache addresses. A conventional implementation first writes dense K and V tensors, then launches another kernel to scatter them into the cache. \fu\ writes them to the cache directly from the QKV epilogue.

For each output row $m$, metadata gives the request and token position $p_m$. The epilogue maps each output column to a head and channel in Q, K, or V. It restores the projection scales, applies RoPE to Q and K, writes Q in the layout used by attention, and stores K and V at

\begin{equation}
\operatorname{offset}(\operatorname{request}(m),\operatorname{head},p_m,\operatorname{channel}).
\end{equation}

Each cache element has one writer, and elements outside the valid range are masked. This removes the dense K/V intermediate tensors and the separate cache-scatter pass.

\subsubsection{Folding the frozen RMSNorm gain}

The block applies RMSNorm before the QKV and gate/up projections. Define

\begin{equation}
q_m=\sum_h x_{mh}^2,
\qquad
\rho_m=\left(q_m/H+\epsilon\right)^{-1/2}.
\end{equation}

For projection weight row $W_n$ and learned gain $\gamma$, the output is

\begin{align}
C_{mn}&=\sum_h x_{mh}\rho_m\gamma_hW_{nh}
=\rho_m\sum_h x_{mh}W'_{nh}, \nonumber\\
W'_{nh}&=\gamma_hW_{nh}. \label{eq:rms_fold}
\end{align}

CODA moves the row factor $\rho_m$ through the next GEMM and applies it in that GEMM's epilogue. During inference, $\gamma$ and $W$ are fixed. \fu\ therefore folds $\gamma$ into $W$ once, before quantizing and repacking the weights. The GEMM then reads the raw activation $X$ and the preprocessed weight $W'$. With FP8 operands, the output is

\begin{equation}
\widetilde C_{mn}=\rho_m a_m b_n\sum_h X^q_{mh}(W')^q_{nh}. \label{eq:rms_fp8}
\end{equation}

The epilogue that produces the residual stream also computes partial sums of squares. A small reduction combines the results from all output tiles to obtain $q_m$. The next projection uses $\rho_m$, so no normalized activation tensor needs to be written to HBM.

\subsubsection{Delayed FP8 emission}

The down projection reads the SwiGLU output $Z$. If the gate/up epilogue writes this output in BF16, a separate quantization kernel must read it and write an FP8 copy. For $M$ rows and feed-forward width $I$, the two approaches move

\begin{equation}
\mathcal B_{\mathrm{separate}}=5MI,
\qquad
\mathcal B_{\mathrm{producer}}=MI, \label{eq:traffic}
\end{equation}

bytes, respectively. The separate path writes BF16, reads it again, and writes FP8. The producer path writes FP8 only once.

\fu\ uses delayed per-tensor scaling. Invocation $t$ quantizes its output with the maximum collected during invocation $t-1$. At the same time, the epilogue computes the current maximum magnitude for invocation $t+1$. It can therefore process and store each fragment immediately. The kernel does not need to keep a full output row in memory on chip while waiting for a global maximum.

The scale state has three requirements. The first invocation must have a valid initial scale. A scale must also be defined for all-zero tensors. Concurrent request groups must keep separate scale states unless they are intentionally run together. Production use also needs saturation counters, because a sudden change in value range can make the previous scale unsuitable.

\subsubsection{Packed prefill}

For requests with lengths $S_1,\ldots,S_B$, padded execution processes $B\max_b S_b$ rows. Packed execution processes only

\begin{equation}
M_{\mathrm{packed}}=\sum_{b=1}^{B}S_b. \label{eq:packed}
\end{equation}

A prefix-sum array maps each packed row to its request and token position. This mapping also defines attention boundaries and KV-cache locations. \fu\ compiles separate kernels for padded and packed inputs, so the padded path does not need runtime layout checks.

\subsection{Fragment-aligned INT4 decode}\label{sec:decode_method}

\subsubsection{Why decode needs a different path}

Let $P$ be the number of projection-weight elements in one layer. Excluding small metadata fields, the minimum number of bytes read during decode is

\begin{equation}
\mathcal B_{\mathrm{decode}}=Pw+2BTH_{kv}Dk, \label{eq:decode_bytes}
\end{equation}

Here, $w$ is the number of bytes per weight, and $k$ is the number of bytes per cached K or V element. For the 8B shape at batch one, the projection weights contain 218.1 million elements. These weights occupy 436 MB in BF16, 218 MB in E4M3, or 109 MB in INT4. Reducing $w$ and $k$ therefore directly reduces the data that must be read and can lower latency.

On H100, the small-$M$ decode kernels cannot use an INT4 MMA that directly reads these packed values. They also have too few rows for the large FP8 WGMMA shapes used in prefill. \fu\ therefore expands low-bit values into 16-bit MMA operands. Its layouts allow this expansion to happen while building the fragments, without a separate dequantization stage.

\subsubsection{INT4 KV-cache mirror}

\fu\ keeps the BF16 KV cache as the main copy for prefill and compatibility, and maintains an INT4 copy for decode. Each byte stores two signed 4-bit values. Each token and KV head has one FP32 scale, calculated as $\operatorname{amax}/7$.

Scaling is performed outside the elementwise loop. For K, the scale is applied to the completed query--key dot product. For V, it is applied to the attention probability before the probability--value MMA. The kernel therefore avoids scaling each unpacked element separately.

A shared-memory table with 256 entries maps each packed byte to two FP16 values. One lookup expands both nibbles into the pair format needed by an MMA fragment. The table does not depend on the token scale, because scales are applied after accumulation.

Keys follow the byte order required by the B operand of \code{mma.sync.m16n8k16}. Each lane reads a contiguous segment the size of its fragment and uses byte-indexed lookups to fill the operand registers directly. This avoids creating a dequantized K tile and moving it through shared memory. V stays in natural channel order because the probability--value MMA groups values differently. It is transposed through shared memory before use.

The attention CTA quantizes the new cache row that it will use in the current decode step. The asynchronous-copy pipeline loads cache scales together with their K/V tile. This keeps cache updates and scale loads outside the serial inner attention loop.

\subsubsection{Hadamard transform for cache outliers}

Uniform INT4 quantization can lose accuracy when a few channels have large values. Following QuaRot and SAW-INT4~\cite{quarot,sawint4}, \fu\ applies a normalized block-diagonal Hadamard matrix $H$ before writing each cache row. The query uses the same transform:

\begin{equation}
(Hq)^{\top}(Hk)=q^{\top}H^{\top}Hk=q^{\top}k.
\end{equation}

The attention output is transformed back once per query row. The transform uses a Walsh--Hadamard butterfly implemented with registers and warp shuffles. Each cache row is rotated once when it is written. The query and output are each rotated once per decode step. The attention loop therefore performs the same work per cached token, and the added rotation cost does not grow with context length. By default, each block covers the full head dimension, but narrower blocks can also be used.

\subsubsection{INT4 projection weights}

Weights are quantized offline with one symmetric scale for each output channel and each group of 128 input elements. The nibbles follow the order required by the B fragments of the decode MMA, so each warp can load a contiguous weight segment for its output tile. The format defines where each nibble is stored, not how its value is chosen. Values can come from the library's round-to-nearest quantizer or a calibrated method such as GPTQ. Both quantization grids are packed into the same layout offline.

Weights are dequantized in registers. Each nibble is inserted into the mantissa bits of the BF16 value 128.0. A logic operation produces packed BF16 values of the form $(128+n,128+m)$. One packed BF16 fused multiply-add then subtracts 136. This removes the base value 128 and the offset of 8 used to convert an unsigned nibble to a signed value. Each group is accumulated separately. Its group scale is applied before the result is added to the running FP32 accumulator. The kernel never creates a full BF16 weight tile.

For $M=1$, each warp reads the weights for one output tile and a quarter of the reduction dimension. The four warps in a CTA combine their partial sums through shared memory. For $2\le M\le32$, a skinny kernel loads activation tiles, packed weights, and scales through a multi-stage asynchronous-copy pipeline. It uses split-K across CTAs when additional CTAs are needed to keep the GPU occupied.

\section{Evaluation}

\subsection{Experimental setup}

We run the experiments on one NVIDIA H100 80GB SXM5 GPU with CUDA 13.0 and PyTorch 2.14.1+cu130. For prefill and CODA measurements, the SM clock is fixed at 1980 MHz. Decode measurements use default clocks. We interleave the compared implementations within the same run so that they experience similar thermal conditions. Unless stated otherwise, compiled PyTorch and \fu\ both use CUDA graphs. Prefill results are the median of four to eight timing invocations after warm-up. For CODA, we alternate the two binaries over three rounds. For decode, all PyTorch and \fu\ candidates run in one process over three interleaved rounds. Both comparisons report median CUDA event time over repeated launches. Each table is a separate interleaved run, so the same configuration can differ by up to 2\% between tables; each comparison is made within one table. The per-layer tests use synthetic inputs and weights, quantized with the library's round-to-nearest quantizer. The whole-model tests in Section~\ref{sec:model} use the trained checkpoint and the GPTQ-packed weights described in Section~\ref{sec:real}. The quantizer changes the nibble values but not their number or storage positions, so it does not change latency. The two quantizers give the same timing within run-to-run variation. The labels 1B, 8B, 70B, and 405B refer to layer shapes. They do not imply that a complete model of that size runs on one GPU.

\begin{table}[t]
\caption{Llama-style layer shapes evaluated in this paper.}
\label{tab:shapes}
\vskip 0.15in
\begin{center}
\begin{footnotesize}
\setlength{\tabcolsep}{4pt}
\begin{tabular}{lrrrrrr}
\toprule
Shape & $H$ & $I$ & $H_q$ & $H_{kv}$ & $D$ & Projection weights \\
\midrule
1B & 2048 & 8192 & 32 & 8 & 64 & 60.82 M \\
8B & 4096 & 14336 & 32 & 8 & 128 & 218.10 M \\
70B & 8192 & 28672 & 64 & 8 & 128 & 855.64 M \\
405B & 16384 & 53248 & 128 & 8 & 128 & 3187.67 M \\
\bottomrule
\end{tabular}
\end{footnotesize}
\end{center}
\vskip -0.1in
\end{table}

The prefill baseline is a compiled, CUDA-graphed PyTorch block using \code{torch.\_scaled\_mm}. It uses one activation scale per row and one weight scale per output channel. The baseline and \fu\ both use FP8 projections and BF16 attention. We calculate effective prefill throughput from the projection FLOPs and causal-attention FLOPs:

\begin{align}
P&=H(H_q+2H_{kv})D+HH_qD+3HI, \\
F_{\mathrm{prefill}}&\approx2BSP+2BS^2H_qD. \label{eq:prefill_flops}
\end{align}

For decode, PyTorch provides an eager BF16 block and a compiled, CUDA-graphed BF16 block. It does not provide a directly comparable full block that combines INT4 weights with an INT4 KV cache. We therefore state the precision of every decode result. We also compare \fu's BF16, FP8, and INT4 paths to measure the effect of reading fewer bytes within the same implementation.

\subsection{Prefill compared with PyTorch}

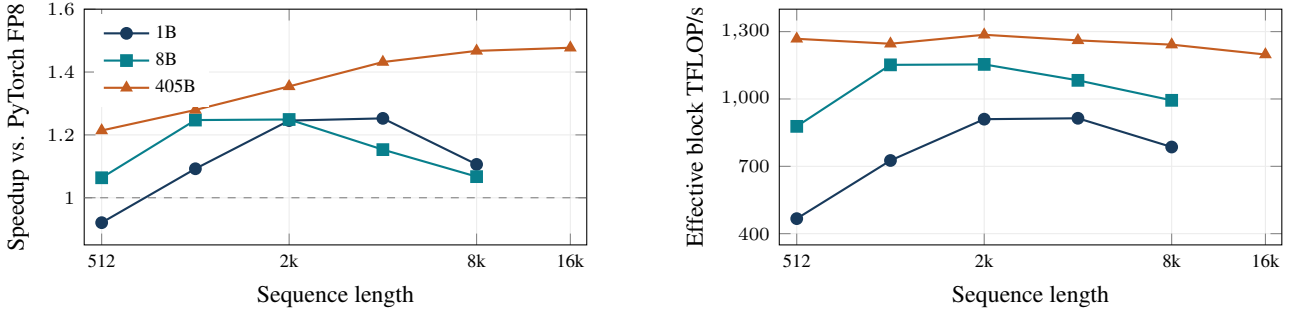
\begin{figure*}[t]
\vskip 0.2in
\begin{center}
\begin{tikzpicture}
\begin{axis}[
  width=0.48\textwidth,height=4.7cm,
  xmode=log,log basis x=2,xmin=450,xmax=18500,
  xtick={512,2048,8192,16384},xticklabels={512,2k,8k,16k},
  xlabel={Sequence length},ylabel={Speedup vs. PyTorch FP8},
  ymin=0.85,ymax=1.6,ytick={1,1.2,1.4,1.6},
  grid=major,grid style={gray!15},
  tick label style={font=\scriptsize},label style={font=\small},
  legend style={font=\scriptsize,at={(0.02,0.98)},anchor=north west,draw=none,fill=white},
  legend cell align=left]
\addplot[ink,thick,mark=*] coordinates {(512,0.92074)(1024,1.09229)(2048,1.24555)(4096,1.25258)(8192,1.10614)};
\addlegendentry{1B}
\addplot[teal,thick,mark=square*] coordinates {(512,1.06362)(1024,1.24721)(2048,1.24901)(4096,1.15314)(8192,1.06719)};
\addlegendentry{8B}
\addplot[orange,thick,mark=triangle*] coordinates {(512,1.21399)(1024,1.27935)(2048,1.35434)(4096,1.43178)(8192,1.46733)(16384,1.47727)};
\addlegendentry{405B}
\addplot[gray,dashed,domain=450:18500] {1};
\end{axis}
\end{tikzpicture}\hfill
\begin{tikzpicture}
\begin{axis}[
  width=0.48\textwidth,height=4.7cm,
  xmode=log,log basis x=2,xmin=450,xmax=18500,
  xtick={512,2048,8192,16384},xticklabels={512,2k,8k,16k},
  xlabel={Sequence length},ylabel={Effective block TFLOP/s},
  ymin=350,ymax=1400,ytick={400,700,1000,1300},
  grid=major,grid style={gray!15},
  tick label style={font=\scriptsize},label style={font=\small}]
\addplot[ink,thick,mark=*] coordinates {(512,467)(1024,726)(2048,910)(4096,914)(8192,786)};
\addplot[teal,thick,mark=square*] coordinates {(512,878)(1024,1152)(2048,1154)(4096,1083)(8192,994)};
\addplot[orange,thick,mark=triangle*] coordinates {(512,1268)(1024,1246)(2048,1286)(4096,1261)(8192,1242)(16384,1198)};
\end{axis}
\end{tikzpicture}
\caption{Prefill at batch one with FP8 projections and BF16 attention. Left: speedup over compiled, CUDA-graphed PyTorch FP8. Right: effective block throughput. Each point shows the measured median.}
\label{fig:prefill}
\end{center}
\vskip -0.2in
\end{figure*}

Figure~\ref{fig:prefill} shows prefill at batch one. Speedup ranges from 0.92\speed\ to 1.48\speed\ across the tested configurations. The 1B shape at $S=512$ is the only measured FP8 case where \fu\ is slower than PyTorch. Avoiding intermediate activation reads and writes becomes more useful for larger shapes or longer sequences. The 405B layer reaches 1286 TFLOP/s at $S=2048$ and stays near 1.20--1.29 PFLOP/s across the measured range.

For the 8B shape at $S=2048$, \fu\ takes 804.2~$\mu$s, compared with 1004.5~$\mu$s for PyTorch, giving a 1.25\speed\ speedup. This agrees with the design goal. The projection mainloop is already close to the library GEMM rate, so most of the improvement comes from avoiding intermediate activation traffic and separate conversion or cache-placement kernels.

\subsection{Closest shared stage compared with CODA}
\label{sec:coda_eval}

CODA targets training and does not implement the same full inference block. A full-block comparison would therefore be misleading. We compare the closest shared operation instead: gate/up projection followed by SwiGLU. CODA runs \code{linear\_swiglu}, and \fu\ runs its gate/up epilogue. Both are tested on the same H100 using \code{coda-kernels} 0.3.1 with \code{nvidia-cutlass-dsl} 4.8.0.

Table~\ref{tab:coda} separates the two comparisons. Panel A is the main precision-matched comparison: compiled PyTorch, CODA, and \fu\ all use BF16. Panel B places the final \fu\ FP8 inference path beside the available BF16 baselines. This compares the final results, but the precisions differ. There is no CODA FP8 row because the released \code{linear\_swiglu} fails an assertion when given FP8 operands on this GPU. Outside SM100 and SM120, its gated output must be 16-bit. In both panels, lower latency is better. Each speedup is the baseline latency divided by the \fu\ latency in the same row.

\begin{table*}[t]
\caption{Gate/up projection followed by SwiGLU at $M=2048$ on one H100. Latency is in $\mu$s; lower is better. Panel A compares BF16 implementations at matched precision. Panel B compares our E4M3 path with BF16 baselines, so the precisions differ. 8B: $K=4096$, $2I=28672$. 405B: $K=16384$, $2I=106496$.}
\label{tab:coda}
\vskip 0.15in
\begin{center}
\begin{small}
\begin{tabular}{lrrrrr}
\toprule
& \multicolumn{3}{c}{Latency ($\mu$s)} & \multicolumn{2}{c}{Our speedup} \\
\cmidrule(lr){2-4}\cmidrule(lr){5-6}
Shape & Compiled PyTorch & CODA & \fu\ (ours) & vs. PyTorch & vs. CODA \\
\midrule
\multicolumn{6}{l}{\textbf{A. Precision matched: all implementations use BF16}} \\
8B & 726 & 690 & \textbf{613} & \textbf{1.18\speed} & \textbf{1.13\speed} \\
405B & 10089 & 9750 & \textbf{9442} & \textbf{1.07\speed} & \textbf{1.03\speed} \\
\addlinespace
\multicolumn{6}{l}{\textbf{B. Cross precision: baselines use BF16; \fu\ uses E4M3}} \\
8B & 726 & 690 & \textbf{316} & \textbf{2.30\speed} & \textbf{2.18\speed} \\
405B & 10089 & 9750 & \textbf{4819} & \textbf{2.09\speed} & \textbf{2.02\speed} \\
\bottomrule
\end{tabular}
\end{small}
\end{center}
\vskip -0.1in
\end{table*}

In the precision-matched BF16 comparison, \fu\ takes 613~$\mu$s for 8B and 9442~$\mu$s for 405B. This is 1.18\speed\ and 1.07\speed\ faster than compiled PyTorch, and 1.13\speed\ and 1.03\speed\ faster than CODA. At 8B, both implementations vary by less than 1\% between rounds. At 405B, \fu\ varies by 4\%, and CODA ranges from 9355--11665~$\mu$s. The 1.03\speed\ result is therefore within the measurement spread and should be treated as similar performance. With E4M3, our final inference path takes 316~$\mu$s and 4819~$\mu$s. The two APIs also perform slightly different work: \fu\ applies the RMSNorm row scale, while CODA saves auxiliary data for backward computation. Table~\ref{tab:coda} therefore compares the closest shared stage, rather than the complete systems.

\subsection{Decode compared with PyTorch}

Table~\ref{tab:decode_torch} compares the INT4 decode block with three PyTorch baselines. Two use BF16: the eager reference block and the same block compiled with CUDA graphs. The third matches the weight precision. It replaces the four projections in the same block with PyTorch's INT4 weight-only kernel, \code{torch.\_weight\_int4pack\_mm}, using group size 128 and BF16 activations. This is the kernel used by gpt-fast and torchao. The INT4 block is compiled and CUDA-graphed in the same way. Normalization, RoPE, cache writes, residuals, and attention use the same code on both sides; only the projection kernels and cache format differ. All candidates run in one process over three interleaved rounds.

When both implementations use a BF16 cache, \fu\ is 1.3--5.0\speed\ faster than the precision-matched INT4 block. This comparison isolates the projection kernels and fusion. With its INT4 cache copy, \fu\ is 1.3--6.3\speed\ faster, and the gap is widest where PyTorch has no matching INT4 cache. The eager INT4 block takes 848--2604~$\mu$s and is limited by launch overhead, as is the eager BF16 block. PyTorch's INT4 kernel is tuned for a single query row, so its advantage over PyTorch's own compiled BF16 block shrinks as rows are added: it is 1.6\speed\ faster at batch one on 8B, 2.2\speed\ faster at batch one on 70B, roughly even at batch 8, and 1.9--2.6\speed\ slower at batch 32. The strongest PyTorch baseline is therefore its INT4 block at small batch and its BF16 block at batch 32, and \fu\ is 1.3--2.4\speed\ faster than whichever is faster in each row.

\begin{table*}[t]
\caption{Full decode-layer latency on one H100, in $\mu$s. All PyTorch baselines use a BF16 cache. The INT4 baseline uses the same block as compiled PyTorch, with its four projections replaced by \code{torch.\_weight\_int4pack\_mm}. Both use the same compilation and CUDA-graph setup. \fu\ uses INT4 weights with either a BF16 cache, matching PyTorch's precision, or its INT4 cache copy. Speedups refer to the complete \fu\ INT4 block.}
\label{tab:decode_torch}
\vskip 0.15in
\begin{center}
\begin{footnotesize}
\setlength{\tabcolsep}{4pt}
\begin{tabular}{lrrrrrrr}
\toprule
& \multicolumn{3}{c}{PyTorch, BF16 cache} & \multicolumn{2}{c}{\fu\ INT4 weights} & \multicolumn{2}{c}{speedup vs.} \\
\cmidrule(lr){2-4}\cmidrule(lr){5-6}\cmidrule(lr){7-8}
Shape, batch, context & eager & compiled & INT4 block & BF16 cache & INT4 cache & compiled & INT4 block \\
\midrule
8B, $B=1$, 1024 & 1015 & 172 & 106 & 77.8 & 78.0 & 2.2\speed & 1.4\speed \\
8B, $B=8$, 1024 & 1019 & 214 & 203 & 76.5 & 74.1 & 2.9\speed & 2.7\speed \\
8B, $B=32$, 4096 & 1118 & 382 & 717 & 265.5 & 180.9 & 2.1\speed & 4.0\speed \\
70B, $B=1$, 1024 & 1399 & 591 & 270 & 205.3 & 206.1 & 2.9\speed & 1.3\speed \\
70B, $B=8$, 1024 & 1405 & 643 & 618 & 196.6 & 197.4 & 3.3\speed & 3.1\speed \\
70B, $B=32$, 4096 & 1498 & 826 & 2138 & 425.4 & 341.5 & 2.4\speed & 6.3\speed \\
\bottomrule
\end{tabular}
\end{footnotesize}
\end{center}
\vskip -0.1in
\end{table*}

Table~\ref{tab:batch} varies batch size while keeping context fixed at 1024, so the two effects can be separated. \fu's INT4 latency stays nearly constant from one to eight sequences because reading the shared weights is still the bottleneck. Latency rises when projection computation becomes the bottleneck. PyTorch's INT4 latency grows roughly linearly, from 106 to 585~$\mu$s, because \code{\_weight\_int4pack\_mm} uses a GEMV-shaped kernel for a single query row. From 16 sequences onward, it is slower than PyTorch's compiled BF16 block, so BF16 becomes the stronger PyTorch baseline. Comparing with the faster PyTorch block at each point, \fu\ achieves speedups of 1.4, 2.7, 2.3 and 2.1\speed\ across this range.

\begin{table}[t]
\caption{Decode-layer latency as batch size increases, with context fixed at 1024 and the 8B shape. Latency is in $\mu$s. The PyTorch INT4 baseline has the same block structure as in Table~\ref{tab:decode_torch}.}
\label{tab:batch}
\vskip 0.15in
\begin{center}
\begin{small}
\begin{tabular}{rrrr}
\toprule
& \multicolumn{2}{c}{PyTorch} & \\
\cmidrule(lr){2-3}
$B$ & compiled & INT4 block & \fu\ INT4 \\
\midrule
1 & 170 & 106 & 77.0 \\
8 & 215 & 202 & 74.6 \\
16 & 229 & 318 & 98.5 \\
32 & 251 & 585 & 118.8 \\
\bottomrule
\end{tabular}
\end{small}
\end{center}
\vskip -0.1in
\end{table}

Table~\ref{tab:decode_precision} tests every combination of projection-weight format (BF16, E4M3, INT4) and KV-cache format (BF16, E4M3 mirror, INT4 mirror) within \fu. This helps separate precision effects from fusion. We use one library build for each cache format and interleave all candidates in one process. The three main decode paths are the diagonal entries: BF16 weights with a BF16 cache, E4M3 weights with an E4M3 cache copy, and INT4 weights with an INT4 cache copy. On this diagonal, INT4 is 31--51\% faster than FP8 and 37--68\% faster than BF16. At batch 8 or below and context 1024, changing the cache format changes latency by at most 2.5\%, within run-to-run variation, because weight traffic dominates. At batch 32 and context 4096, the E4M3 cache saves 55--61~$\mu$s and the INT4 cache 86--89~$\mu$s, for every weight format. The gain is largest when projection-weight traffic dominates.

\begin{table*}[t]
\caption{Decode-layer latency within \fu\ under CUDA graph replay, in $\mu$s. Column groups specify the KV-cache format, and columns within each group specify the projection-weight format. INT4 cache results include rotation. The three main decode paths are BF16/BF16, E4M3/E4M3 and INT4/INT4.}
\label{tab:decode_precision}
\vskip 0.15in
\begin{center}
\begin{small}
\setlength{\tabcolsep}{4pt}
\begin{tabular}{lrrrrrrrrr}
\toprule
& \multicolumn{3}{c}{BF16 cache} & \multicolumn{3}{c}{E4M3 cache} & \multicolumn{3}{c}{INT4 cache} \\
\cmidrule(lr){2-4}\cmidrule(lr){5-7}\cmidrule(lr){8-10}
Shape, batch, context & BF16 & FP8 & INT4 & BF16 & FP8 & INT4 & BF16 & FP8 & INT4 \\
\midrule
8B, $B=1$, 1024 & 165.2 & 122.1 & 77.0 & 167.2 & 124.4 & 78.0 & 166.9 & 122.4 & 77.0 \\
8B, $B=8$, 1024 & 174.8 & 128.4 & 76.4 & 176.1 & 128.6 & 78.3 & 174.0 & 126.2 & 75.4 \\
8B, $B=32$, 4096 & 364.6 & 324.4 & 266.2 & 306.0 & 269.9 & 205.5 & 275.3 & 236.8 & 179.8 \\
70B, $B=1$, 1024 & 588.7 & 334.4 & 205.0 & 590.6 & 335.6 & 206.2 & 603.3 & 334.4 & 206.8 \\
70B, $B=8$, 1024 & 606.6 & 399.9 & 196.1 & 607.0 & 398.8 & 199.8 & 600.8 & 398.2 & 195.2 \\
1B, $B=1$, 1024 & 60.5 & 51.9 & 39.2 & 63.5 & 55.1 & 39.7 & 61.8 & 53.4 & 38.2 \\
\bottomrule
\end{tabular}
\end{small}
\end{center}
\vskip -0.1in
\end{table*}

These speedups come from three sources: fewer weight bytes, fewer KV-cache bytes, and block fusion. The comparison with PyTorch therefore does not measure fusion alone. The within-\fu\ table separates the first two effects. Lower weight precision reduces latency for every shape. Lower cache precision reduces latency when cache reads make up a large share of the total bytes.

\subsection{Whole-model decode and prefill}\label{sec:model}

Table~\ref{tab:model} evaluates the complete Llama-3.1-8B model. Every decoder layer uses the fused block, while PyTorch runs the embedding, final RMSNorm, and LM head. We capture one greedy decode step in a single CUDA graph and replay it. The step takes a token as input and produces the next token. Its timing includes the 32 block launches and reads of the 1~GB BF16 LM-head weights, which are unchanged by the decode-weight format. The PyTorch baselines use the transformers implementation of the same BF16 checkpoint with a static cache, in eager mode or compiled with CUDA graphs. We time decode steps at positions just below the reported context length, using 64 steps per round for five rounds and reporting the median. Prefill uses one 2048-token prompt at batch one. INT4 is supported only for decode, so the INT4 rows use BF16 prefill.

\begin{table*}[t]
\caption{Complete Llama-3.1-8B on one H100: latency and throughput per decode step, and prefill of a 2048-token prompt. Each \fu\ row lists the projection-weight format / KV-cache format used by its decode blocks.}
\label{tab:model}
\vskip 0.15in
\begin{center}
\begin{footnotesize}
\setlength{\tabcolsep}{4pt}
\begin{tabular}{lrrrrrrrr}
\toprule
& \multicolumn{3}{c}{Decode, ms per step} & \multicolumn{3}{c}{Decode, tokens/s} & \multicolumn{2}{c}{Prefill 2048} \\
\cmidrule(lr){2-4}\cmidrule(lr){5-7}\cmidrule(lr){8-9}
Implementation & $B=1$, 1024 & $B=8$, 1024 & $B=32$, 4096 & $B=1$ & $B=8$ & $B=32$ & ms & tokens/s \\
\midrule
PyTorch eager, BF16 & 22.77 & 22.30 & 103.74 & 44 & 359 & 309 & 67.9 & 30\,147 \\
PyTorch compiled, BF16 & 8.08 & 11.08 & 59.35 & 124 & 722 & 539 & 52.1 & 39\,318 \\
\fu\ BF16 / BF16 & 5.51 & 5.77 & 11.71 & 182 & 1\,386 & 2\,732 & 46.4 & 44\,153 \\
\fu\ FP8 / BF16 & 4.19 & 4.33 & 10.56 & 239 & 1\,849 & 3\,030 & 26.6 & 77\,010 \\
\fu\ INT4 / BF16 & 2.70 & 2.66 & 8.76 & 371 & 3\,004 & 3\,654 & 46.5 & 44\,067 \\
\fu\ BF16 / INT4 & 5.62 & 5.79 & 8.98 & 178 & 1\,383 & 3\,562 & 44.7 & 45\,859 \\
\fu\ FP8 / INT4 & 4.25 & 4.40 & 7.99 & 235 & 1\,817 & 4\,006 & 27.8 & 73\,550 \\
\fu\ INT4 / INT4 & 2.76 & 2.64 & 6.02 & 362 & 3\,030 & 5\,313 & 46.6 & 43\,913 \\
\bottomrule
\end{tabular}
\end{footnotesize}
\end{center}
\vskip -0.1in
\end{table*}

The complete INT4 model decodes 2.0--2.2\speed\ faster than \fu's BF16 model at batch 8 or below, and 1.9\speed\ faster at batch 32 and context 4096. In the latter case, INT4 cache storage contributes about as much as INT4 weights. A step takes 11.7~ms with BF16 weights and cache, 8.8 with INT4 weights alone, and 6.0 with both. Compared with \fu's BF16 path, the whole-model gain is smaller than the per-layer gain in Table~\ref{tab:decode_torch}. At batch one, the embedding and LM head add about 0.3~ms to every path. Prefilling 2048 tokens takes 46~ms with BF16 and 27~ms with FP8, compared with 52~ms for compiled PyTorch.

The whole-model speedup over the transformers implementation is 2.9--9.9\speed, larger than the 2.1--2.9\speed\ in Table~\ref{tab:decode_torch}. Profiling the compiled PyTorch step explains this difference. At batch 32 and context 4096, attention accounts for 89\% of its 60.8~ms step. Copy kernels take 29~ms to expand the 8 KV heads to the 32 query heads and arrange the cache for cuDNN. The cuDNN attention kernel takes another 23~ms to process the full 4096-entry static cache. Projections take 5.9~ms, close to 32 times the per-layer block. At batch one, projection GEMMs account for 81\% of the step, and weight traffic explains the performance gap. The transformers static-cache attention path therefore takes about ten times as long as PyTorch SDPA on the compact cache used in the per-layer benchmark. \fu\ reads its 8-head INT4 cache directly in fragment order, without expanding the heads. This explains most of the whole-model advantage. Table~\ref{tab:model} uses BF16 PyTorch rows because transformers has no equivalent whole-model INT4 path available in this setup. The precision-matched kernel result is the per-layer comparison in Table~\ref{tab:decode_torch}; the whole-model ratio also includes these framework-level differences.

\subsection{Design ablation}\label{sec:ablation}

Table~\ref{tab:ablation} uses compile-time switches to replace design elements of the INT4 decode block with earlier versions, while keeping precision and data size fixed. Storing K in natural order and transposing it through shared memory, rather than using the MMA fragment order in Section~\ref{sec:decode_method}, increases latency by 16\% at batch 32 and context 4096. It costs at most 2.3\% where cache reads account for only a small share of the bytes. If per-nibble arithmetic also replaces the byte-indexed lookup, the combined latency increase is 25\%. These two changes reproduce the earlier cache design. Quantizing the new cache row in a separate kernel between QKV and attention, rather than inside attention, adds 1--2~$\mu$s per step. Converting weight nibbles with an extra bias MMA, instead of register logic and FMA operations, increases latency by 9\% at batch one, 20\% at batch 8, and 11\% on 70B. The earlier FMA split-KV attention kernel takes 9--17\% more time than the Tensor Core INT4-cache kernel at context 1024, and 2.4\speed\ as long at batch 32 and context 4096. The earlier skinny BF16 projection kernel does not support low-bit formats, so we compare it only on the BF16 path. For the three 8B shapes, its latencies are 192, 232 and 299~$\mu$s, compared with 167, 174 and 276 for the current kernel.

\begin{table}[t]
\caption{Design ablation of the INT4 decode block using INT4 weights and an INT4 cache. Latency is in $\mu$s. Each variant replaces a design element with its earlier version.}
\label{tab:ablation}
\vskip 0.15in
\begin{center}
\begin{footnotesize}
\setlength{\tabcolsep}{3pt}
\begin{tabularx}{\columnwidth}{Yrrrr}
\toprule
& \multicolumn{3}{c}{8B} & 70B \\
\cmidrule(lr){2-4}\cmidrule(lr){5-5}
Variant & $B=1$ & $B=8$ & $B=32$, 4096 & $B=1$ \\
\midrule
\fu\ as described & 77.3 & 74.8 & 178.3 & 208.1 \\
K in natural order, shared-memory transpose & 77.9 & 76.5 & 207.0 & 207.5 \\
\quad and nibbles unpacked arithmetically & 78.1 & 78.2 & 223.0 & 208.8 \\
Cache row quantized by a separate kernel & 79.4 & 76.7 & 179.4 & 209.5 \\
Weight nibbles converted by a bias MMA & 84.2 & 89.8 & 179.8 & 231.4 \\
FMA split-KV attention kernel & 84.0 & 87.3 & 423.4 & 210.1 \\
\bottomrule
\end{tabularx}
\end{footnotesize}
\end{center}
\vskip -0.1in
\end{table}

\subsection{Numerical results}

We compare single-layer outputs with two references. Prefill tests use a separate FP64 implementation, and decode tests use the PyTorch BF16 block. These tests measure numerical error on synthetic data. Section~\ref{sec:real} evaluates the INT4 weight path on a trained checkpoint.

Table~\ref{tab:prefill_accuracy} reports prefill RMSE at $S=512$. On these synthetic inputs, the delayed-scale FP8 path has 1.8--2.0\% higher RMSE than the PyTorch FP8 baseline with row-wise scales. The fused BF16 path has lower error. This is consistent with performing several operations between BF16 GEMMs in FP32, but this experiment does not measure that effect separately.

\begin{table}[t]
\caption{Absolute RMSE of the block output compared with an FP64 reference at $S=512$. Values are in units of $10^{-3}$.}
\label{tab:prefill_accuracy}
\vskip 0.15in
\begin{center}
\begin{footnotesize}
\setlength{\tabcolsep}{4pt}
\begin{tabularx}{\columnwidth}{Yrrrr}
\toprule
& \multicolumn{2}{c}{Normal input} & \multicolumn{2}{c}{Outlier mixture} \\
\cmidrule(lr){2-3}\cmidrule(lr){4-5}
Configuration & 1B & 8B & 1B & 8B \\
\midrule
PyTorch BF16 & 4.13 & 4.12 & 4.20 & 4.18 \\
\fu\ BF16 & 3.46 & 3.43 & 3.54 & 3.49 \\
PyTorch FP8, row-wise scale & 39.6 & 39.5 & 39.1 & 39.1 \\
\fu\ FP8, delayed tensor scale & 40.3 & 40.3 & 39.8 & 39.8 \\
\bottomrule
\end{tabularx}
\end{footnotesize}
\end{center}
\vskip -0.1in
\end{table}

Table~\ref{tab:decode_accuracy} summarizes decode error. On random data, the INT4 cache has about three times the RMS error of the E4M3 cache. INT4 weights follow the same trend. Increasing four K channels per head by 8\speed\ sharply raises the unrotated INT4 cache error. The Hadamard transform reduces this error from 22.2\% to 7.4\% for the 8B shape, and from 29.9\% to 12.9\% for the 1B shape.

\begin{table}[t]
\caption{Relative RMS output error for one decode layer, in \%. Cache results use a PyTorch BF16 reference. Weight results isolate weight quantization while keeping the cache unchanged. All inputs are synthetic.}
\label{tab:decode_accuracy}
\vskip 0.15in
\begin{center}
\begin{small}
\begin{tabularx}{\columnwidth}{Yrr}
\toprule
Configuration & 8B & 1B \\
\midrule
E4M3 cache & 0.9 & 0.7 \\
INT4 cache & 3.2 & 2.2 \\
INT4 cache, channel outliers, no rotation & 22.2 & 29.9 \\
INT4 cache, channel outliers, Hadamard rotation & 7.4 & 12.9 \\
\addlinespace[2pt]
E4M3 weights & 3.4--3.5 & 3.4--3.6 \\
INT4 weights, group size 128 & 9.8--10.2 & 9.9--10.5 \\
\bottomrule
\end{tabularx}
\end{small}
\end{center}
\vskip -0.1in
\end{table}

The remaining INT4 error comes from quantization rather than a mismatch in the kernel implementation. The next subsection measures this error on a trained checkpoint and shows that GPTQ reduces it without changing the packed execution format. Under channel outliers, the rotated INT4 cache is much more accurate than the unrotated cache, although E4M3 remains more accurate.

\subsection{INT4 decode on a trained checkpoint}\label{sec:real}

The synthetic tests isolate kernel error. To measure the quality cost of the INT4 decode path on a real model, we evaluate the BF16 release of Llama-3.1-8B~\cite{llama3} on the WikiText-2 test set~\cite{wikitext}, following the GPTQ protocol~\cite{gptq}. We concatenate the test split and divide it into 141 non-overlapping windows of 2048 tokens. Perplexity is the exponential of the mean token negative log-likelihood. Every token is decoded one position at a time through the production decode block of each of the 32 layers. PyTorch runs the embedding, final RMSNorm, and LM head, and 32 windows are decoded in parallel. Table~\ref{tab:real_ppl} uses a BF16 KV cache to isolate weight quantization. Table~\ref{tab:real_full} uses the INT4 cache copy and evaluates the complete INT4 decode block. GPTQ calibration uses 32 random windows of 2048 tokens from the WikiText-2 training split. GPTQ uses symmetric 4-bit weights, static groups of 128, no column reordering, 5\% damping, and an FP64 error-feedback solve on the GPU. Using the WikiText-2 validation split, we selected round-to-nearest quantization for layer 1's down projection, where Llama-3 forms the massive activations that later layers attend to~\cite{massive}; the test split was used only to produce the reported numbers. An offline packer stores the GPTQ grid in the kernel layout and matches the library quantizer byte for byte. The kernels themselves are unchanged.

Table~\ref{tab:real_ppl} shows the results. The BF16 kernel path matches PyTorch perplexity to three decimal places. For round-to-nearest weights, the INT4 kernel path and a BF16 simulated-quantization reference built from the same grid agree to 0.001 perplexity. Round-to-nearest INT4 weights increase perplexity by 0.99, while GPTQ weights in the same format increase it by 0.75. For this model, choosing each group's scale with a mean-squared-error search, rather than the group maximum, makes round-to-nearest quantization less accurate.

\begin{table}[t]
\caption{WikiText-2 perplexity for Llama-3.1-8B. Every token runs through the fused decode blocks with a BF16 KV cache. The test uses 141 windows of 2048 tokens.}
\label{tab:real_ppl}
\vskip 0.15in
\begin{center}
\begin{small}
\begin{tabularx}{\columnwidth}{Yrr}
\toprule
Decode-block weights & Perplexity & vs.\ BF16 \\
\midrule
PyTorch BF16 (reference) & 6.240 & \\
\fu\ BF16 kernels & 6.240 & +0.000 \\
\fu\ INT4, round-to-nearest & 7.226 & +0.986 \\
\fu\ INT4, round-to-nearest, clipped scales & 7.393 & +1.153 \\
\fu\ INT4, GPTQ & 6.986 & +0.746 \\
\bottomrule
\end{tabularx}
\end{small}
\end{center}
\vskip -0.1in
\end{table}

Table~\ref{tab:real_full} adds the INT4 KV-cache copy. Its last row therefore evaluates the complete INT4 decode block. With Hadamard rotation, the cache adds 0.15--0.23 perplexity beyond the weight-quantization cost, which is less than the cost of the weights themselves. The complete block with GPTQ weights reaches 7.18, or 0.94 above BF16. Without rotation, quantizing the cache alone adds 0.54, and the complete block reaches 7.76. Rotation therefore removes about 70\% of the cache's quality cost on the trained model. This agrees with the synthetic outlier results in Table~\ref{tab:decode_accuracy}.

\begin{table}[t]
\caption{WikiText-2 perplexity for Llama-3.1-8B using the fused decode blocks and an INT4 KV-cache copy, with and without Hadamard rotation. The test uses 141 windows of 2048 tokens. The BF16-cache column repeats Table~\ref{tab:real_ppl}.}
\label{tab:real_full}
\vskip 0.15in
\begin{center}
\begin{small}
\begin{tabularx}{\columnwidth}{Yrrr}
\toprule
& & \multicolumn{2}{c}{INT4 cache mirror} \\
\cmidrule(lr){3-4}
Decode-block weights & BF16 cache & rotated & unrotated \\
\midrule
BF16 & 6.240 & 6.387 & 6.777 \\
INT4, round-to-nearest & 7.226 & 7.458 & 8.161 \\
INT4, GPTQ & 6.986 & 7.182 & 7.762 \\
\bottomrule
\end{tabularx}
\end{small}
\end{center}
\vskip -0.1in
\end{table}

Table~\ref{tab:real_layer} measures per-layer error in two ways. The independent tests evaluate one layer at a time, using the hidden state entering that layer in the BF16 model. We prefill 512 real tokens with BF16 weights, then decode four more real tokens with the block's INT4 projections. The reference uses the same input and block with BF16 weights. BF16 kernels differ from an FP32 reference by 0.3--0.4\%. Round-to-nearest INT4 adds 10--13\% relative RMS error to a layer's decode output, while GPTQ adds 4.5--7\%. The cascaded tests decode 16 windows through the INT4 stack using its own residual stream and KV cache. The error after each layer therefore includes errors from earlier layers and token positions.

\begin{table}[t]
\caption{Relative RMS error of each layer's decode output, in \%, using real Llama-3.1-8B weights and activations. Independent: each layer receives the BF16 model's input to that layer; results are averaged over four decode tokens after a 512-token prefill. Cascaded: the INT4 stack decodes 16 windows using its own residual stream and KV cache; results cover positions 1024--2047. RTN means round-to-nearest. The last column compares the BF16 kernels with an FP32 reference.}
\label{tab:real_layer}
\vskip 0.15in
\begin{center}
\begin{footnotesize}
\setlength{\tabcolsep}{4pt}
\begin{tabularx}{\columnwidth}{Yrrrrr}
\toprule
& \multicolumn{2}{c}{Independent} & \multicolumn{2}{c}{Cascaded} & BF16 \\
\cmidrule(lr){2-3}\cmidrule(lr){4-5}
Layer & RTN & GPTQ & RTN & GPTQ & kernels \\
\midrule
0 & 13.0 & 6.1 & 15.6 & 7.6 & 0.28 \\
1 & 12.6 & 7.2 & 20.9 & 13.1 & 0.39 \\
2 & -- & -- & 30.7 & 21.4 & -- \\
4 & -- & -- & 36.5 & 28.4 & -- \\
8 & -- & -- & 40.1 & 29.3 & -- \\
11 & -- & -- & 41.6 & 30.4 & -- \\
15 & 10.1 & 6.2 & 37.3 & 28.2 & 0.29 \\
24 & -- & -- & 31.7 & 24.5 & -- \\
31 & 11.6 & 4.5 & 36.3 & 25.8 & 0.27 \\
\bottomrule
\end{tabularx}
\end{footnotesize}
\end{center}
\vskip -0.1in
\end{table}

After layer 31, cascaded error is 36\% with round-to-nearest weights and 26\% with GPTQ. It reaches 42\% and 30\%, respectively, near layer 11, then decreases. Errors are nearly the same for positions 256--1023 and 1024--2047. For the output logits, KL divergence from the BF16 next-token distribution is 0.14 and 0.08 nats per token, respectively. The top-1 prediction matches BF16 for 83.5\% and 88\% of tokens.

\subsection{The served path and downstream tasks}\label{sec:prod}

Tables~\ref{tab:real_ppl} and~\ref{tab:real_full} decode every token from the start of each window. This isolates the decode path, but differs from the serving workflow. Table~\ref{tab:prod} evaluates that workflow. The first 1024 tokens of each window use BF16 or E4M3 prefill, since INT4 is supported only for decode. The remaining 1023 positions are decoded one at a time. We score only the decoded positions, and the cache copy contains all positions.

In the serving workflow, the complete INT4 decode block adds 0.53 perplexity, or 9.4\% over the all-BF16 path. When every token is decoded from the first position, the increases are 0.94 and 15.1\%. The protocols differ in two ways. In the serving test, each scored token has at least 1024 tokens of context, and the first half of the window is processed with higher-precision weights. FP8 prefill adds only 0.004 perplexity, and E4M3 decode weights add 0.02. Thus, the INT4 decode weights account for the full quality cost with either prefill path.

\begin{table}[t]
\caption{WikiText-2 perplexity for Llama-3.1-8B using the serving workflow. The first 1024 tokens use prefill, and the remaining 1023 are decoded one at a time. Only decoded positions are scored, over 141 windows. Prefill does not support INT4. A dash indicates an unmeasured combination.}
\label{tab:prod}
\vskip 0.15in
\begin{center}
\begin{small}
\begin{tabularx}{\columnwidth}{Yrr}
\toprule
Prefill / decode weights & BF16 cache & INT4 cache \\
\midrule
BF16 / BF16 & 5.595 & 5.728 \\
BF16 / E4M3 & 5.616 & -- \\
BF16 / INT4 round-to-nearest & 6.158 & -- \\
BF16 / INT4 GPTQ & 5.986 & 6.120 \\
E4M3 / BF16 & 5.599 & 5.731 \\
E4M3 / E4M3 & 5.621 & 5.745 \\
E4M3 / INT4 GPTQ & 5.989 & 6.123 \\
\bottomrule
\end{tabularx}
\end{small}
\end{center}
\vskip -0.1in
\end{table}

Perplexity does not directly measure task accuracy, so Table~\ref{tab:tasks} adds two multiple-choice benchmarks. These evaluations score each answer option with a single forward pass rather than a decode loop, so they cannot use the decode kernels directly. We instead use a PyTorch model carrying the values the kernels compute with. The packed weights are dequantized and returned to the checkpoint's original layout, and hooks quantize K and V for each token and head as the cache copy stores them, with one $\operatorname{amax}/7$ scale per token and head and K rotated first. For the weights this is the same simulated quantization that agrees with the kernel path to 0.001 perplexity; the cache hooks follow the kernel's definition but were not separately cross-checked at this model size. INT4 weights reduce accuracy by 2.1 percentage points on HellaSwag and 4.9 on ARC-Challenge. The INT4 cache adds losses of 0.4 and 2.7 points, respectively. Repeating the two INT4 configurations independently changes the results by at most 0.3 points, within the standard errors of 0.4 and 1.5 points. ARC-Challenge shows a larger quality loss than perplexity alone suggests. The format increases perplexity by 15\% in the all-decode test and 9\% in the serving test, and reduces accuracy by up to 7.6 points on the harder of the two tasks.

\begin{table}[t]
\caption{Zero-shot accuracy (\code{acc\_norm}) of Llama-3.1-8B with the decode path's quantization. The evaluation model uses the same numerical values as the kernels. Standard errors are 0.004 for HellaSwag and 0.015 for ARC-Challenge.}
\label{tab:tasks}
\vskip 0.15in
\begin{center}
\begin{small}
\begin{tabularx}{\columnwidth}{Yrr}
\toprule
Configuration & HellaSwag & ARC-Challenge \\
\midrule
BF16 & 0.793 & 0.552 \\
INT4 GPTQ weights, BF16 cache & 0.772 & 0.503 \\
\quad and the rotated INT4 cache & 0.768 & 0.476 \\
\bottomrule
\end{tabularx}
\end{small}
\end{center}
\vskip -0.1in
\end{table}

\subsection{Variable-length prefill}

Table~\ref{tab:packed} compares packed and padded execution within \fu\ for the 8B shape, with $B=8$ and $S_{\max}=2048$. For equal-length inputs, packing adds no visible overhead. When one request has length 2048 and the other seven have length 128, packed execution is 5.07\speed\ faster because it skips most padded projection rows.

\begin{table}[t]
\caption{Packed and padded prefill compared within the same \fu\ implementation.}
\label{tab:packed}
\vskip 0.15in
\begin{center}
\begin{footnotesize}
\setlength{\tabcolsep}{4pt}
\begin{tabularx}{\columnwidth}{Yrrr}
\toprule
Length distribution & Actual tokens & Padding ratio & Speedup \\
\midrule
All equal & 16384 & 1.00\speed & 1.00\speed \\
$U(0.7,1.0)\cdot S_{\max}$ & 14097 & 1.16\speed & 1.15\speed \\
$U(0.2,1.0)\cdot S_{\max}$ & 10448 & 1.57\speed & 1.57\speed \\
Lognormal, median $S_{\max}/3$ & 8527 & 1.92\speed & 1.85\speed \\
One at $S_{\max}$, seven at $S_{\max}/16$ & 2944 & 5.57\speed & 5.07\speed \\
\bottomrule
\end{tabularx}
\end{footnotesize}
\end{center}
\vskip -0.1in
\end{table}

\section{Discussion and Limitations}

\paragraph{Baseline precision.}
The PyTorch prefill comparison uses the same projection and attention precisions, but different activation-scaling policies. The BF16 gate/up comparison between \fu\ and CODA is the closest like-for-like result, although their training and inference APIs keep different auxiliary data. The comparison between INT4 decode and the PyTorch BF16 reference uses different precisions: the reference has BF16 weights and a BF16 cache. Comparing BF16, FP8, and INT4 within \fu\ isolates the effect of storage precision.

\paragraph{Scope.}
Performance tests cover a single Transformer layer on one H100 and the complete Llama-3.1-8B model in Section~\ref{sec:model}. The quality tests in Section~\ref{sec:real} use the same complete model. These tests do not include multi-layer scheduling, distributed communication, request admission, or complete serving latency. The 405B label describes one layer's dimensions; we do not run a full 405B model on one GPU.

\paragraph{Accuracy.}
Synthetic tests cover one layer. Trained-checkpoint tests cover one model, one dataset for perplexity, and two zero-shot multiple-choice tasks for accuracy. INT4 decode increases perplexity by 9--15\%, depending on the protocol, and reduces ARC-Challenge accuracy by up to 7.6 percentage points. This is a meaningful quality loss. Delayed FP8 scaling is tested on stable workloads after warm-up. Production use also needs safe scale initialization, saturation monitoring, and request-local state. Task accuracy on trained checkpoints must still be validated before deployment.

\paragraph{Portability.}
The implementation is designed for Hopper fragment layouts, asynchronous copies, and resource limits. Supporting another GPU generation requires different tile shapes, pipelines, and packed layouts. The main principles can still be applied, but the kernel details depend on the architecture.

\section{Conclusion}

\fu\ optimizes prefill and decode separately. Prefill keeps FP8 GEMM throughput high and reduces data movement between operations. It extends CODA's GEMM-plus-epilogue formulation with inference-specific cache placement, fixed RMSNorm-gain folding, delayed FP8 producer output, and packed inputs. Decode reads fewer bytes by using INT4 weights and an INT4 KV-cache copy. Both formats follow the layout required by the consuming MMA fragments, so values are dequantized close to where the Tensor Cores use them.

On one H100, \fu\ prefill is up to 1.48\speed\ faster than compiled PyTorch FP8 and reaches 1.29 PFLOP/s. For the closest shared BF16 gate/up--SwiGLU stage, \fu\ is 1.13\speed\ faster than CODA at the 8B shape. Over the 8B and 70B shapes at batch 1 to 32, the INT4 decode block is 1.3--2.4\speed\ faster than the stronger of PyTorch's two decode blocks at each point. Measured separately, it is 2.1--3.3\speed\ faster than the compiled, CUDA-graphed BF16 block and 1.3--6.3\speed\ faster than the same block using PyTorch's INT4 projection kernel. It is also 31--51\% faster than \fu's FP8 path. The complete 8B model decodes 2.9\speed\ faster than compiled PyTorch at batch one. On Llama-3.1-8B, INT4 weights increase WikiText-2 perplexity by 0.99 with round-to-nearest quantization or 0.75 with GPTQ in the same format. The rotated INT4 cache adds 0.20, giving a complete INT4 block that is 0.94 above BF16. Without rotation, quantizing the cache alone would add 0.54. The kernel path matches a simulated-quantization reference to within 0.001 perplexity. These results support GEMM epilogues for prefill and fragment-aligned low-bit execution for decode.

\bibliography{futransformer}
\bibliographystyle{mlsys2025}

\end{document}